\documentclass[letterpaper, 10 pt, conference]{ieeeconf}  % Comment this line out
\IEEEoverridecommandlockouts                              % This command is only
\usepackage{xcolor}
\usepackage{amsmath,amsfonts,amssymb}
\usepackage{graphicx}
\usepackage{optidef}
\usepackage{dsfont}

\DeclareMathOperator*{\argmax}{arg\,max}
\DeclareMathOperator*{\argmin}{arg\,min}
\newcommand{\mc}[1]{\mathcal{#1}}
\newcommand{\mf}[1]{\mathfrak{#1}}

\title{\LARGE \bf
Value of Information-based Deceptive Path Planning \\ Under Adversarial Interventions
}

\author{Wesley A. Suttle, Jesse Milzman, Mustafa O. Karabag, Brian M. Sadler, Ufuk Topcu % <-this % stops a space
\thanks{W. A. Suttle and J. Milzman are with the U.S. Army Research Laboratory, Adelphi, MD 20783
        {\tt\small \{wesley.a.suttle.ctr, jesse.milzman.civ\}@army.mil}. M. O. Karabag, B. M. Sadler, and U. Topcu are with the University of Texas at Austin, Austin, TX 78712 {\tt\small \{karabag, utopcu\}@utexas.edu, brian.sadler@austin.utexas.edu }.} %
}

\begin{document}

\maketitle
\thispagestyle{empty}
\pagestyle{empty}

%%%%%%%%%%%%%%%%%%%%%%%%%%%%%%%%%%%%%%%%%%%%%%%%%%%%%%%%%%%%%%%%%%%%%%%%%%%%%%%%
\begin{abstract}

Existing methods for deceptive path planning (DPP) address the problem of designing paths that conceal their true goal from a passive, external observer. Such methods do not apply to problems where the observer has the ability to perform adversarial interventions to impede the path planning agent. In this paper, we propose a novel Markov decision process (MDP)-based model for the DPP problem under adversarial interventions and develop new value of information (VoI) objectives to guide the design of DPP policies.
Using the VoI objectives we propose, path planning agents deceive the adversarial observer into choosing suboptimal interventions by selecting trajectories that are of low informational value to the observer. Leveraging connections to the linear programming theory for MDPs, we derive computationally efficient solution methods for synthesizing policies for performing DPP under adversarial interventions. In our experiments, we illustrate the effectiveness of the proposed solution method in achieving deceptiveness under adversarial interventions and demonstrate the superior performance of our approach to both existing DPP methods and conservative path planning approaches on illustrative gridworld problems.

\end{abstract}
%%%%%%%%%%%%%%%%%%%%%%%%%%%%%%%%%%%%%%%%%%%%%%%%%%%%%%%%%%%%%%%%%%%%%%%%%%%%%%%%

\section{Introduction} \label{sec:introduction}

Deceptive path planning (DPP) is the problem of designing a path that conceals its true objective from an outside observer. Several approaches to this problem have recently been developed, using model-based planning \cite{masters2017deceptive, ornik2018deception, karabag2021deception, chen2024deceptive} and model-free reinforcement learning \cite{liu2021deceptive,lewis2023deceptive,fatemi2024deceptive,kim2025deceptive}. These methods make the strong assumption that the observer is passive and unable to affect the path planning agent's environment, however, significantly limiting their applicability.
%
% \todo{describe approaches accounting for adversarial observers, highlight their intractability; state that middle ground combining tractability with adversarial interventions is required}

In this paper, we consider the problem of performing DPP in settings where the observer has the ability to perform adversarial interventions to impede the path planning agent. While conservative approaches using zero-sum games \cite{HART199219} or robust Markov Decision Processes (MDPs) \cite{NE:05} can produce feasible paths to the desired goal under worst-case adversarial interventions, to our knowledge, the problem of leveraging deception and influencing the observer's decisions to achieve better-than-worst-case performance under adversarial interventions remains unaddressed. The challenges to addressing this issue are three-fold: (i) the lack of a concrete model for DPP that incorporates adversarial interventions; (ii) the lack of principled deception metrics capturing the impact of deceptiveness on interventions and agent path costs; (iii) the lack of practical solution procedures for obtaining DPP policies in this setting.

We address these challenges as follows: we propose a novel MDP-based model for DPP under adversarial interventions, we formulate two novel value of information (VoI)-based deception metrics that capture the impact of deceptive policies through the informational value of the beliefs they induce in the observer, and we derive tractable approximations of our proposed VoI deception metrics and develop efficient, linear programming (LP)-based solution methods for the resulting VoI DPP problems. To evaluate the effectiveness of the proposed approach, we provide qualitative and quantitative experimental comparisons with the passive-observer methods of \cite{savas2022deceptive} and a conservative path planning (CPP) baseline on a variety of gridworld problems. Our qualitative results indicate that VoI DPP provides flexible levels of deceptiveness in the proposed model, while passive-observer methods are ill-suited to the adversarial setting and CPP provides only inflexible, worst-case solutions. Our quantitative results demonstrate that, under actual observer interventions, our VoI DPP methods leverage deception to achieve lower-cost paths compared to both passive-observer and CPP approaches when interventions occur during the \textit{critical deception window} of the DPP problem.

% Deceptive path planning (DPP) is the problem of designing a path that conceals its true objective from an outside observer. Existing methods for DPP \cite{fatemi2024deceptive, savas2022deceptive, masters2017deceptive, ornik2018deception} suffer from two main drawbacks: (i) the notions of deception used are overly simplistic, failing to capture more realistic notions of deceptiveness, and (ii) the observer is unable to affect the environment, ruling out more realistic scenarios where the observer can take adversarial actions. To address these issues, in this project we propose a general-purpose Value-of-Information (VoI)-based notion of deception to tackle issue (i), while we take an important step towards addressing issue (ii) by considering an adversarial observer with the ability to block the agent's path.
% %
% In the context of deceptive path planning (DPP) as considered in \cite{fatemi2024deceptive, savas2022deceptive, masters2017deceptive, ornik2018deception}, this project proposes a general, Value-of-Information (VoI)-based notion of deception to be used in formulating the DPP problem. This VoI-based formulation strictly generalizes but also captures existing notions of deception. We consider specific applications of this framework to designing paths that deceive an adversary that has the ability to place obstacles and set traps for the planner.

\subsection{Related work}

Formal study of deceptive path planning began with \cite{masters2017deceptive}, while \cite{ornik2018deception} proposed an MDP formulation and LP-based solutions for general deceptive planning. Following this, \cite{karabag2019optimal} studied deception for supervisory control, \cite{chen2024deceptive} addressed deception for resource allocation, and \cite{savas2022deceptive} considered the DPP problem. Most relevant for our work is \cite{savas2022deceptive}, which combines an observer prediction model based on the principle of maximum entropy \cite{ziebart2009planning, ziebart2010modeling} with \cite{ornik2018deception} to obtain optimal DPP policies. We emphasize that all DPP methods mentioned above are \textit{passive-observer} in the sense that the observer is unable to interfere with the path planning agent's environment. This assumption limits their applicability, motivating us to pursue their extension to the adversarial setting. We note that \cite{cates2023planning} provides tools for deceptive planning from the observer perspective for a specific attacker-entrapment problem, but the problem setting and notion of deception considered are restrictive, contrasting with the general setting considered in our work. For further discussion of \cite{cates2023planning} and active-observer approaches beyond DPP, see Section \ref{app:related_works}.

The literature on goal obfuscation and goal recognition is closely related to the deceptive planning setting we consider. The goal obfuscation works \cite{kulkarni2018resource, bernardini2020optimization} considered the problem of generating paths over a graph that conceal the agent's true objective from a passive observer. The paper \cite{takahashi2024transit} addressed the transit obfuscation problem, an extension of goal obfuscation to concealing an intermediate ``transit'' point that must be visited en route to the agent's goal from a passive observer. Goal recognition works including \cite{masters2021extended, price2023domain} address the problem of modeling observer inference of the agent's true goal. While \cite{masters2021extended} focuses primarily on observer modeling, \cite{price2023domain} proposes corresponding deception metrics and formulates deceptive planning approaches. We emphasize that the goal obfuscation and recognition works discussed above focus on passive-observer settings.

% \textbf{Active-observer deception.} While previous methods for DPP are restricted to the passive-observer setting, there exists previous work on deceptive planning under adversarial interventions and deception in games. The paper \cite{cates2023planning} considered the problem of deceptive planning for a defender attempting to guide an attacker into a set of trap states while concealing its intentions from the attacker. There are two key differences between the setting of \cite{cates2023planning} and that considered in our paper: (i) our approach addresses the attacker-/agent-perspective problem of previous works \cite{ornik2018deception, savas2022deceptive, chen2024deceptive, fatemi2024deceptive}, while \cite{cates2023planning} considers the distinct defender-/observer-perspective deceptive planning problem; (ii) the notion of deception considered in \cite{cates2023planning} is based on a binary hypothesis test of whether the attacker realizes it is being deceived, while in our work we develop general VoI deception metrics capturing the value of arbitrary observer belief distributions.

The notion of VoI originates from early work in statistical decision theory \cite{howard1966information} and comparison of experiments \cite{blackwell1953equivalent}. While we are not aware of any works explicitly using VoI for deception, \cite{greenberg1982role} noted that misleading manipulation of an opponent's information to one's advantage is considered deceptive -- for zero-sum games, this is equivalent minimizing the opponent's VoI. In this paper, we study VoI-based objectives within the context of DPP.
For additional related works, see Section \ref{app:related_works}.

\section{Background} \label{sec:background}

In this section we provide the necessary background on the LP-based approach to DPP proposed in \cite{savas2022deceptive}.
%
% Its key components are: a weighted graph over which the underlying path planning is to be performed, a corresponding MDP, an observer model defined relative to the MDP, deception formulations defined with respect to the observer model, and LP-based solution methods for obtaining deceptive policies for the resulting DPP problem.
%
% We will discuss each of these components in this section, as they form the basis of the extension that we propose in Section \ref{sec:formulation} below.

\subsection{Markov decision processes and graph abstractions} \label{subsec:background:graph_MDP}

Consider an undirected, weighted graph $\mf{G} = (\mc{S}, \mc{A}, c)$, where $\mc{S}$ denotes the set of nodes, $\mc{A} \subset \mc{S} \times \mc{S}$ the set of edges, and $c : \mc{A} \rightarrow \mathbb{R}$ the edge weights. Let $\mc{A}(s) \subset \mc{A}$ denote the set of edges having node $s \in \mc{S}$ as as endpoint.
Consider next a corresponding discounted Markov decision process (MDP) $\mc{M} = (\mc{S}, \mc{A}, p, c, s_0, \gamma)$, where the states $\mc{S}$ are simply the nodes of the graph $\mf{G}$, the actions $\mc{A}$ are the edges of $\mc{G}$ and $\mc{A}(s) \subset \mc{A}$ denotes the available actions at state $s$, $p : \mc{S} \times \mc{A} \rightarrow \Delta(\mc{S})$ is the transition probability function mapping state-action pairs to probability distributions over $\mc{S}$, $s_0 \in \mc{S}$ is the start state, $\gamma \in (0, 1)$ is the discount factor, and we slightly abuse notation by letting $c(s, a) = c(a)$, where $c(a)$ is the edge weight associated with $a$ in $\mf{G}$. Note that, in the graph-based setting we consider, $p(s' | s, a) > 0$ for only those $s' \in \mc{S}$ such that $a = (s, s')$ or $a = (s', s)$, i.e., only when $s, s'$ are adjacent.
In the sequential decision-making problem formalized by this MDP, an agent interacts with the system using a policy $\pi : \mc{S} \rightarrow \Delta(\mc{A})$ mapping states to probability distributions over $\mc{A}$. Note that $\pi(a | s) = 0$ when $a \notin \mc{A}(s)$. At a given timestep $t \in \mathbb{N}$, the agent is in state $s_t$, samples $a \sim \pi(\cdot | s_t)$, receives cost $c_t = c(s_t, a_t)$, and transitions to a new state $s_{t+1} \sim p(\cdot | s_t, a_t)$.
To model the path planning component of our problem, let $\mc{G} \subset \mc{S}$ denote the set of candidate goals available to the agent and let $G^* \in \mc{G}$ denote the agent's true goal. Next, let $P^{\pi}(\text{Reach}(G))$ denote the probability that following policy $\pi$ over the MDP $\mc{M}$ eventually leads the agent into state $G \in \mc{G}$, and let $R_{max}(G) = \max_{\pi} P(\text{Reach}(G))$ denote the maximum probability of reaching $G$ under any policy over $\mc{M}$. Note that, for connected graphs $\mf{G}$ and suitable conditions (e.g., irreducibility and ergodicity) on $p$, we will have $R_{max}(G) = 1$, for all $G \in \mc{G}$. A natural goal-seeking objective in this setting would be to find a policy $\pi^*$ such that $R_{max}(G)$ is maximized while achieving minimal cost, i.e., such that $\pi^* = \argmin_{\pi} J(\pi)$, where $J(\pi) := \mathbb{E}_{\pi} \left[ \sum_{t=0}^\infty \gamma^t c_t \right]$. Finally, let $\zeta = (s_0, a_0, s_1, a_1, \ldots)$ denote a possible trajectory of state-action pairs in $\mc{M}$, and, for a given $T \in \mathbb{N}$, let $\zeta_{0:T} = (s_0, a_0, \ldots, s_T, a_T)$ denote the induced partial trajectory up to time $T$.

\subsection{Observer model} \label{subsec:background:observer_model}

In order to arrive at a meaningful notion of deceptiveness, we first need a tractable model of how the observer expects the path planning agent to behave within the MDP $\mc{M}$, given a specific goal $G \in \mc{G}$, as well as a model of the observer's beliefs regarding the identity of the agent's true goal $G^* \in \mc{G}$. For the former, a natural approach is to assume that the observer expects them to approximately follow the minimum cost path to $G$, as captured in \eqref{opt:obs_model} below. For the latter, once we are equipped with our model of the expectations of the observer, this can be used to model the observer's beliefs regarding the agent's true goal given observed trajectory information.
%
% For the latter, \cite{savas2022deceptive} provides an observer belief model that captures the observer's beliefs regarding the agent's true goal given partial trajectory information. This is achieved by leveraging \cite{ziebart2009planning, ziebart2010modeling} to compute a belief distribution $P(G | \zeta_{0:T})$ over all $G \in \mc{G}$, conditioned on the agent's trajectory $\zeta_{0:T}$ up to time $T$. This can be accomplished using the approximation
%
These are simultaneously achieved in \cite{savas2022deceptive} by leveraging the approach developed in \cite{ziebart2009planning, ziebart2010modeling} to compute a belief distribution $P(G | \zeta_{0:T})$ over all $G \in \mc{G}$, conditioned on the agent's trajectory $\zeta_{0:T}$ up to time $T$. Specifically, \cite{savas2022deceptive} uses the approximation
\begin{equation}
    P(G | \zeta_{0:T}) \approx \widehat{P}(G | s_T) := \frac{ e^{V_G(s_T) - V_G(s_0)} P(G) }{ \sum_{G' \in \mc{G}} e^{V_{G'}(s_T) - V_{G'}(s_0)} P(G') } \label{eqn:observer_model}
\end{equation}
for the observer model, where $P(\cdot)$ encodes the observer's prior beliefs regarding the identity of the agent's true goal and $V_G$ is the softmax value function corresponding to the optimal policy $\pi_G$ for the following problem over $\mc{M}$:
\begin{mini}
  {\pi}{ \mathbb{E}_{\pi} \left[ \sum_{t=0}^\infty \gamma^t \left( c_t - \alpha H(\pi(\cdot | \ s_t)) \right) \right] }{}{}
  \addConstraint{ P_{\pi}( \text{Reach}(G) ) }{= R_{max}(G), }{} \label{opt:obs_model}
\end{mini}
where $H(\pi(\cdot | s)) = - \sum_{a} \pi(a | s) \log \pi(a | s)$ is the policy entropy of policy $\pi$, $P_{\pi}(\text{Reach}(G))$ is the probability of eventually reaching $G$ within $\mc{M}$ while following $\pi$, and $\alpha > 0$ is a scalar parameter capturing the level of inefficiency the observer expects the agent to demonstrate.
It is shown in \cite{ziebart2009planning, ziebart2010modeling} that $V_G$ satisfying the equations
\begin{align}
    Q_G(s, a) &= - c(s, a) + \gamma \sum_{s' \in \mc{S}} p(s' | s, a) V_G(s') \label{eqn:softmax_Q} \\
    V_G(s) &= \alpha \log \sum_{a \in \mc{A}} e^{Q_G(s, a) / \alpha} \label{eqn:softmax_V}
\end{align}
can be computed by performing softmax value iteration $|\mc{G}|$ times, providing an explicit method to obtain the belief distribution in  \eqref{eqn:observer_model}.

\subsection{Deception formulations} \label{subsec:background:deception}

Equipped with the observer model detailed above, we next formally describe the two common types of deception considered in the DPP literature: exaggeration and ambiguity. We here provide the definitions from \cite{savas2022deceptive} (see \cite{masters2017deceptive, ornik2018deception} for additional discussion). Intuitively, a deceptive agent \textit{exaggerates} by navigating towards a decoy goal instead of the true goal. Formally, this can be captured by the cost
\begin{equation}
    f(s_t, a_t) = 1 + P(G^* | \zeta_{0:t} ) - \max_{G \in \mc{G} \setminus G^*} P(G | \zeta_{0:t} ). \label{eqn:classical_exaggeration}
\end{equation}
To see why \eqref{eqn:classical_exaggeration} can be used to encourage the exaggerated behavior, notice that, if we take $\pi$ to be the policy used to generate $\zeta_{0:t}$, equation \eqref{eqn:classical_exaggeration} achieves its minimum when $\pi$ is such that $P(G^* | \zeta_{0:t} ) = 0$ and $\max_{G \in \mc{G} \setminus G^*} P(G | \zeta_{0:t} ) = 1$, and is otherwise increasing in $P(G^* | \zeta_{0:t} ) - \max_{G \in \mc{G} \setminus G^*} P(G | \zeta_{0:t} )$. While minimizing the expectation of \eqref{eqn:classical_exaggeration} leads to completely deceptive behavior, constraining our search to minimize \eqref{eqn:classical_exaggeration} subject to $G^*$ eventually being reached yields paths that exaggerate the likelihood of entering decoy goal states by passing close by them, yet that still reach the true goal.

A deceptive agent demonstrates \textit{ambiguous} behavior, on the other other hand, by generating paths that remain noncommittal for as long as possible before turning towards the true goal. This is represented formally via the function
\begin{equation}
    \small
    f(s_t, a_t) = 
    \begin{cases}
        \sum_{G, G' \in \mc{G}} | P(G | \zeta_{0:t}) - P(G' | \zeta_{0:t}) | &\text{ if } s_t \in \mc{S} \setminus \mc{G} \\
        0 &\text{ o.w.}
    \end{cases}
    \label{eqn:classical_ambiguity}
\end{equation}
Policies minimizing \eqref{eqn:classical_ambiguity} in expectation will tend to assign all $g \in \mc{G}$ equal probability, implying that a policy generating such trajectories will render the true goal $G^*$ indistinguishable to the observer from the other elements of $\mc{G}$.

\subsection{Linear programming-based solution} \label{subsec:background:lp_solution}

% \todo{move this discussion to \S\ref{sec:propsed_approach}}

We next describe the linear programming (LP) solution procedures proposed in \cite{savas2022deceptive} for obtaining policies to perform DPP using the objectives \eqref{eqn:classical_exaggeration}, \eqref{eqn:classical_ambiguity} defined above. Let a given graph-based MDP $\mathcal{M}$, set of candidate goals $\mathcal{G}$, and true goal $G^* \in \mc{G}$ be given. Equipped with a deceptive cost function $g : \mathcal{S} \times \mathcal{A} \rightarrow \mathbb{R}$, a natural approach is to search for a policy $\pi^*$ that minimizes long-run expected cost subject to the probability of reaching $G^*$ under $\pi^*$ being as large as possible within $\mathcal{M}$, i.e., to solve the problem
\begin{mini}
  {\pi}{ \mathbb{E}_{\pi} \left[ \sum_{t=0}^\infty g(s_t, a_t) \right] }{}{}
  \addConstraint{ P_{\pi}( \text{Reach}(G^*) ) }{= R_{max}(G^*), }{} \label{opt:dpp_0}
\end{mini}

To render \eqref{opt:dpp_0} tractable, two things are needed: a cost $g$ that simultaneously captures the desired notion of deception and ensures that the objective of \eqref{opt:dpp_0} is well-defined; an explicit formulation of the constraint in \eqref{opt:dpp_0} in a mathematically tractable form (e.g., as a system of linear equality constraints). To achieve the first, \cite{savas2022deceptive} proposes $g(s, a) := \gamma_a^{T_{min}(s)} f(s, a)$, where $\gamma_a \in (0, 1]$ is an ancillary discount factor that can be tuned to control trajectory length and $T_{min}(s) := d_{\mf{G}}(s_0, s)$, where $d_{\mf{G}}(s', s'')$ denotes the shortest path from $s'$ to $s''$ in the graph $\mf{G}$. Equipped with this cost function, \cite{savas2022deceptive} proposes the following linear program over the space of state-action occupancy measures
%
% $\Delta(\mc{S} \times \mc{A}) = \{ \lambda \in \mathbb{R}^{|\mc{S}|\cdot|\mc{A}|} \ | \ \lambda \geq 0 \text{ and } \lambda^T \mathds{1} = 1 \}$,
%
$\Delta(\mc{S} \times \mc{A}) = \{ \lambda \in \mathbb{R}^{|\mc{S}|\cdot|\mc{A}|} \ | \ \lambda \geq 0 \}$,
which modifies the classical dual LP formulation for solving MDPs (see, e.g., \cite{Puterman_2014, altman2021constrained}) to include a constraint corresponding to that in \eqref{opt:dpp_0}:
\small
\begin{mini}
  {\lambda \geq 0}{ \sum_{ s \in \mc{S}, a \in \mc{A} } g(s, a) \lambda(s, a) }{}{}
  \addConstraint{ \sum_{a \in \mc{A}} \lambda(s, a) - \hspace{-4mm} \sum_{s' \in \mc{S}, a \in \mc{A} } \hspace{-2mm} p(s' | s, a) \lambda(s', a) }{= \mathds{1}_{ \{s_0\} }(s),}{\hspace{-2mm} \forall s }{} \label{opt:dpp_1}
  \addConstraint{ \sum_{ s \in \mc{S}, a \in \mc{A} } \lambda(s, a) r(s, a) }{= R_{max}(G^*)}{}{}
  %
  % \addConstraint{ \sum_{ s \in \mc{S}, a \in \mc{A} } \lambda(s, a)}{= 1,}{}{}
\end{mini}
\normalsize
where $\mathds{1}_{B}(\cdot)$ is the indicator function on the set $B$ and $r(s,a) := p(G^* | s, a)$. It is well-known (again, see \cite{Puterman_2014, altman2021constrained}) that, given an optimal state-action occupancy measure $\lambda^*$ of \eqref{opt:dpp_1}, a corresponding optimal policy can be recovered via $\pi^*(a | s) = \lambda^*(s, a) / \sum_{a' \in \mc{A}} \lambda^*(s, a')$, and, more generally, that whenever $\pi^{\lambda}(a | s) = \lambda(s, a) / \sum_{a' \in \mc{A}} \lambda(s, a')$, then $\pi^{\lambda}$ induces the occupancy measure $\lambda$ over $\mc{M}$. In the problem \eqref{opt:dpp_1}, the linear constraint $\sum_{s, a} \lambda(s, a) r(s, a) = R_{max}(G^*)$ ensures that any feasible state-action occupancy measure $\lambda$ yields a policy $\pi^{\lambda}$ such that the reachability constraint of problem \eqref{opt:dpp_0} is satisfied.

Though solving \eqref{opt:dpp_1} yields the desired optimal deceptive policy, it is observed in \cite{savas2022deceptive} that the resulting policy may visit zero-cost states too frequently, as this does not affect the optimal objective function value. To find an optimal policy that reaches $G^*$ as quickly as possible while still achieving the optimal value $v^*$ of the deception objective in \eqref{opt:dpp_1}, \cite{savas2022deceptive} proposes performing an additional search over the set of optimal policies by solving the problem

{
\small
\begin{mini}
    {\lambda \geq 0}{ \sum_{s \in \mc{S}} \sum_{a \in \mc{A}} \lambda(s, a) }{}{}
    \addConstraint{ \sum_{a \in \mc{A}} \lambda(s, a) - \hspace{-4mm} \sum_{s' \in \mc{S}, a \in \mc{A} } \hspace{-2mm} p(s' | s, a) \lambda(s', a) }{= \mathds{1}_{ \{s_0\} }(s),}{\hspace{0mm} \forall s }{}
    \addConstraint{ \sum_{ s \in \mc{S}, a \in \mc{A} } \lambda(s, a) r(s, a) }{= R_{max}(G^*)}{}{}
    \addConstraint{ \sum_{s \in \mc{S}} \sum_{a \in \mc{A}} g(s, a) \lambda(s, a) }{= v^*}{} \label{opt:dpp_2}
    %
    % \addConstraint{ \sum_{s \in \mc{S}} \sum_{a \in \mc{A}} \lambda(s, a)}{= 1.}{}
y \end{mini}
\normalsize
}
Having obtained an optimal state-action occupancy measure $\lambda^*$ by solution of \eqref{opt:dpp_2}, we can recover our final, optimal DPP policy via $\pi^*(a | s) = \lambda^*(s, a) / \sum_{a' \in \mc{A}} \lambda^*(s, a')$.
% \section{Problem Formulation} \label{sec:formulation}
\section{Modeling Adversarial Interventions} \label{sec:formulation}

% In this section we develop a VoI-based extension of the classical approach presented in Section \ref{sec:background} and provide corresponding solution methods. Our extension generalizes existing DPP methods in two key ways: first, whereas previous approaches consider problems where the observer is external to and unable to affect the agent, we consider a new, adversarial problem setting where the observer has the ability to intervene to change the underlying graph and its objective is to select the intervention that results in the greatest increase in cost for the agent to reach its true goal; second, we propose a novel VoI-based formulation of deceptiveness and develop a corresponding objective that can be minimized to obtain policies for performing DPP in our new, adversarial problem setting. Despite the increased complexity of this adversarial, VoI-based DPP problem, we derive an efficient, LP-based solution approach for obtaining DPP policies.

In this section, we propose a generalization of the passive-observer approach presented in Section \ref{sec:background} that incorporates adversarial interventions.
%
% In this setting the observer has the ability to alter the underlying graph to adversely impact the agent.
%
% In this new setting, the observer has the ability to intervene to change the underlying graph with the objective of selecting the intervention that results in the greatest increase in cost to the agent according to the the observer's beliefs regarding the agent's true goal.
% %
% % As a baseline against which to compare the VoI-based approaches proposed in the following section,
% %
% % As a useful baseline, we also propose a conservative path planning (CPP) method based on an extensive form game formulation \cite{HART199219} that generates feasible paths for the agent to the true goal under worst-case observer interventions.
%
%
%
% % \subsection{Planning under adversarial interventions} \label{subsec:formulation:adversarial_setting}
%
%
%
% We now present the adversarial setting considered in this paper.
%
In this new setting, the observer has the ability to affect the agent by carrying out interventions that change the underlying graph, where each intervention consists of removing certain nodes and edges from the original graph and altering the edge weights of the induced subgraph. Given a candidate goal for the agent, each candidate intervention will alter the cost to the agent of navigating to that goal within the resulting subgraph and induced MDP. The observer aims to select the intervention that increases the agent's navigation costs the most, according to its current beliefs regarding the agent's true goal. In the remainder of this section, we provide technical details of the problem setting and demonstrate how navigation costs can be explicitly computed for given goals and candidate interventions. 
%The machinery established here will be critical to the VoI-based DPP framework developed in subsequent sections.

Recall the undirected, weighted graph $\mf{G} = (\mc{S}, \mc{A}, c)$ and associated MDP $\mc{M} = (\mc{S}, \mc{A}, p, c, s_0, \gamma)$ of Section \ref{subsec:background:graph_MDP}. Let an intervention be denoted by $(\bar{\mc{S}}, \bar{\mc{A}}, \bar{c})$, where $\bar{\mc{S}} \subset \mc{S}$ is such that $s_0 \cup \mc{G} \subset \bar{\mc{S}}$, $\bar{\mc{A}} \subset \mc{A}$, and $\bar{c} : \bar{\mc{A}} \rightarrow \mathbb{R}$. Notice that $(\bar{\mc{S}}, \bar{\mc{A}})$ is a subgraph of the unweighted graph underlying $\mf{G}$, and that the intervention $(\bar{\mc{S}}, \bar{\mc{A}}, \bar{c})$ is simply this subgraph equipped with a new edge weighting. Let $\mc{I} = \{ (\bar{\mc{S}}^i, \bar{\mc{A}}^i, \bar{c}^i) \}_{i=1, \ldots, k}$ denote a set of $k$ candidate interventions. For the $i$th candidate intervention we will abuse notation and directly write $i = (\bar{\mc{S}}^i, \bar{\mc{A}}^i, \bar{c}^i)$. For intervention $i$, define $\bar{\mc{M}}^i = (\bar{\mc{S}}^i, \bar{\mc{A}}^i, \bar{p}^i, \bar{c}^i, s_0, \gamma)$ to be the associated MDP, where we further abuse notion as in Section \ref{subsec:background:graph_MDP} by writing $\bar{c}^i(s, a) = \bar{c}^i(a)$, and where $\bar{p}^i$ is the transition probability function induced from $p$ by the selection of intervention $i$.
\footnote{
In general, for all $s, s' \in \mc{S}$ and $a \in \mc{A}$ we will have $\bar{p}^i(s' | s, a) = p(s' | s, a)$, except when $s \notin \bar{\mc{S}}^i$, $s' \notin \bar{\mc{S}}^i$, or $a \notin \bar{\mc{A}}^i$. When $s \notin \bar{\mc{S}}^i$ or $a \notin \bar{\mc{A}}^i$, the probability distribution $p( \cdot | s, a)$ is no longer meaningful and can be discarded; when $s \in \bar{\mc{S}}^i$ and $a \in \bar{\mc{A}}^i$ but $s' \notin \bar{\mc{S}}^i$, any positive probability mass assigned to $p(s' | s, a)$ needs to be appropriately redistributed over $\bar{\mc{S}}^i$ to ensure that $\sum_{s'' \in \bar{\mc{S}}^i } \bar{p}^i( s'' | s, a) = 1$. In the simplest case this can be achieved by setting $\bar{p}^i(s | s, a) = p(s | s, a) + p(s' | s, a)$, but in the general case this redistribution can be problem-dependent.
}
For given candidates $G \in \mc{G}$ and $i \in \mc{I}$, an optimal policy for attaining $G$ in the induced MDP $\bar{\mc{M}}^i$ can be obtained by solving the problem
\small
\begin{mini}
  {\pi}{ \mathbb{E}_{\pi, \bar{\mc{M}}^i } \left[ \sum_{t=0}^\infty \gamma^t \left( \bar{c}^i(s_t, a_t) - \alpha H(\pi(\cdot | s_t)) \right) \right] }{}{}
  \addConstraint{ \bar{P}^i_{\pi}( \text{Reach}(G) ) }{= \bar{R}^i_{max}(G), }{} \label{opt:induced_opt_0}
\end{mini}
\normalsize
where the expectation in the objective is being taken over the Markov chain induced by $\pi$ in the MDP $\bar{\mc{M}}^i$, $\bar{P}^i_{\pi}$ denotes the probability of eventually reaching $G$ in $\bar{\mc{M}}^i$ by following $\pi$, and $\bar{R}^i_{max}(G) = \max_{\pi} \bar{P}^i_{\pi}(G)$ denotes the maximum probability of reaching $G$ in $\bar{\mc{M}}^i$ under any policy. When the underlying MDP $\mc{M}$ is clear from the context, we will denote the solution to \eqref{opt:induced_opt_0} by $\bar{\pi}^i_G$, and the corresponding optimal objective function value by $J(\bar{\pi}^i_G)$. We abuse terminology here and refer to \textit{the} optimal policy, though there may be multiple.
%
% \footnote{There may be multiple optimal solutions to \eqref{opt:induced_opt_0}. Since we are primarily concerned with the corresponding optimal objective function value, which is unique, however, we refer to solution policies in the singular for convenience.}
%
Practically, the optimal value $J(\bar{\pi}^i_G)$ can be obtained by performing softmax value iteration using equations \eqref{eqn:softmax_Q}-\eqref{eqn:softmax_V} to obtain the optimal softmax state value function $\bar{V}^i_G$ for problem \eqref{opt:induced_opt_0}. Given start state $s_0$ as well as goal $G \in \mc{G}$ and intervention $i \in \mc{I}$, the optimal value in \eqref{opt:induced_opt_0} is then given by $J(\bar{\pi}^i_G) = \bar{V}^i_G(s_0)$. The set of induced optima $\{ J(\bar{\pi}^i_G) \}_{(i, G) \in \mc{I} \times \mc{G}}$ corresponding to the interventions available to the observer under the different candidate goals will be key to the VoI objectives developed in Section \ref{sec:proposed_approach}.

The foregoing provides a computationally efficient way for the observer to evaluate the impact of candidate interventions on an agent navigating to any candidate goal. When combined with the observer model of Section \ref{sec:background}, it also equips the agent with tools to perform deceptive path planning, which we turn to next.

\section{Deceptive Path Planning Under \\ Adversarial Interventions} \label{sec:proposed_approach}

In this section, we propose novel VoI objectives for DPP, formulate the VoI-based DPP problem, and provide LP-based solution procedures. We first formalize the notion of the information available to the observer as well as the value of that information in the DPP context, then propose corresponding VoI metrics of deceptiveness based on the set $\{ J(\bar{\pi}^i_G) \}_{(i, G) \in \mc{I} \times \mc{G}}$ of intervention-induced costs obtained from problem \eqref{opt:induced_opt_0}. We then recast these deception metrics as tractable optimization objectives for DPP and pose the VoI-based DPP problem as an LP. This problem can be efficiently solved to obtain globally optimal solutions to the VoI DPP problem with only a minor increase in computational expense over the method of Section \ref{sec:background}: in the VoI DPP setting we perform softmax value iteration $|\mc{I}| \cdot |\mc{G}|$ times before solving the LP, whereas in the classical setting we perform softmax value iteration $|\mc{G}|$ times before solving the LP.

\subsection{Value of Information deception objectives} \label{subsec:formulation:voi_defs}

Let $\Delta(\mc{G})$ denote the set of probability distributions over the set $\mc{G}$ of candidate goals. In the problem setting under consideration, we take each element $b \in \Delta(\mc{G})$ to represent the observer's belief distribution regarding the identity of the agent's true goal given the information available to the observer. Given an element $b \in \Delta(\mc{G})$ and leveraging the machinery developed in Section \ref{sec:formulation}, the expected value to the observer of selecting intervention $i \in \mc{I}$ is given by
\begin{equation}
    \mathbb{E}_b \left[ J(\bar{\pi}^i_G) \right] = \sum_{G \in \mc{G}} b(G) J(\bar{\pi}^i_G). \label{eqn:expected_removal_cost_wrt_b}
\end{equation}

The goal of the observer is to select, based on the information available to it, the intervention that most increases the navigation costs to the agent.
For a given belief $b \in \Delta(\mc{G})$, we therefore define the \textit{observer value of belief} to be
%
% \begin{equation}
%     \text{VoB}(b) = \max_{i \in \mc{I}} \mathbb{E}_b \left[ J(\bar{\pi}^i_G) \right] = \max_{i \in \mc{I}} \sum_{G \in \mc{G}} b(G) J(\bar{\pi}^i_G), \label{eqn:vob_def}
% \end{equation}
%
\begin{equation}
    \text{VoB}^{o}(b) = \max_{i \in \mc{I}} \sum_{G \in \mc{G}} b(G) J(\bar{\pi}^i_G), \label{eqn:vob_o_def}
\end{equation}
the expected highest intervention cost that the observer can impose on the agent under its belief $b$. Since we are ultimately interested in evaluating the observer's beliefs and actions from the perspective of the agent, we also consider the alternative \textit{agent value of belief} defined by
%
% \begin{equation}
%     \text{VoB}^{a}(b) = J(\bar{\pi}^{i^*}_{G^*}), \label{eqn:vob_a_def}
% \end{equation}
% %
% where
% %
% \begin{equation}
%     i^* = \argmax_{i \in \mc{I}} \sum_{G \in \mc{G}} b(G) J(\bar{\pi}^i_G) \label{eqn:bar_s_def}
% \end{equation}
%
\begin{align}
    \text{VoB}^{a}(b) &= J(\bar{\pi}^{i^*}_{G^*}), \label{eqn:vob_a_def} \\
    i^* &= \argmax_{i \in \mc{I}} \sum_{G \in \mc{G}} b(G) J(\bar{\pi}^i_G), \label{eqn:bar_s_def}
\end{align}
and $\bar{\pi}^{i^*}_{G^*}$ is the optimal policy for the true goal $G^*$ in the induced MDP $\bar{\mc{M}}^{i^*}$. Equation \eqref{eqn:vob_a_def} captures the expected cost to the agent of the observer behavior that the belief $b$ will induce. As we will see in Sections \ref{subsubsec:formulation:vob_o}-\ref{subsubsec:formulation:vob_a} below, \eqref{eqn:vob_o_def} and \eqref{eqn:vob_a_def} intuitively correspond to the ambiguity and exaggeration notions of deception, respectively.

In order to render the value of belief expressions defined in \eqref{eqn:vob_o_def} and \eqref{eqn:vob_a_def} practically useful, we need a tractable representation of the belief $b$. To this end, we make the natural assumption that the information available to the observer at time $T$ is the partial trajectory, $\zeta_{0:T}$, generated by the agent up to time $T$. Given $\zeta_{0:T}$, a natural approach to supplying the belief distribution $b \in \Delta(\mc{G})$ induced by this information is to leverage the observer model described in Section \ref{subsec:background:observer_model}. Under this model, given information $\zeta_{0:T}$ we can directly invoke equation \eqref{eqn:observer_model} to obtain 
\begin{equation}
    \footnotesize
    b(G) = P(G | \zeta_{0:T}) \approx \widehat{P}(G | s_T) = \frac{ e^{V_G(s_T) - V_G(s_0)} P(G) }{ \sum_{G' \in \mc{G}} e^{V_{G'}(s_T) - V_{G'}(s_0)} P(G') }. \label{eqn:belief_model}
\end{equation}
Substituting \eqref{eqn:belief_model} into \eqref{eqn:vob_o_def} yields the tractable approximation
\begin{subequations}
\begin{align}
    \text{VoB}^o(\zeta_{0:T}) &= \max_{i \in \mc{I}} \sum_{G \in \mc{G}} P(G | \zeta_{0:T}) J(\bar{\pi}^i_G) \label{eqn:vob_approx_0} \\
    &\approx \max_{i \in \mc{I}} \sum_{G \in \mc{G}} \widehat{P}(G | s_T) J(\bar{\pi}^i_G), \label{eqn:vob_approx_1}
\end{align}
%
% \begin{align}
%     \small
%     \text{VoB}^o(\zeta_{0:T}) = \max_{i \in \mc{I}} \sum_{G \in \mc{G}} P(G | \zeta_{0:T}) J(\bar{\pi}^i_G) \approx \max_{i \in \mc{I}} \sum_{G \in \mc{G}} \widehat{P}(G | s_T) J(\bar{\pi}^i_G), \label{eqn:vob_approx_1}
% \end{align}
\end{subequations}
where the $J(\cdot)$, $V_G(\cdot)$, and $P(\cdot)$ values are obtained as in Sections \ref{subsec:background:observer_model} and \ref{sec:formulation}. Similarly, substituting \eqref{eqn:belief_model} into \eqref{eqn:bar_s_def},
\begin{equation}
    \widehat{i^*} = \argmax_{i \in \mc{I}} \sum_{G \in \mc{G}} \widehat{P}(G | s_T) J(\bar{\pi}^i_G), \label{eqn:hat_s_def}
\end{equation}
where $\widehat{i^*}$ approximates $i^*$ from \eqref{eqn:bar_s_def}. Using \eqref{eqn:hat_s_def} in \eqref{eqn:vob_a_def} yields
\begin{equation}
    \text{VoB}^a(\zeta_{0:T}) \approx J(\bar{\pi}^{\widehat{i^*}}_{G^*}). \label{eqn:vob_a_approx_0}
\end{equation}
Since for a fixed start state $s_0$ the approximations of $\text{VoB}^o(\zeta_{0:T})$ and $\text{VoB}^a(\zeta_{0:T})$ given in \eqref{eqn:vob_approx_1} and \eqref{eqn:vob_a_approx_0} depend on $\zeta_{0:T}$ only through the final state $s_T$, in what follows we will use the shorthand $\text{VoB}^o(s_T)$ and $\text{VoB}^a(s_T)$ to denote $\text{VoB}^o(\zeta_{0:T})$ and $\text{VoB}^a(\zeta_{0:T})$, respectively.

Now that we are equipped with computationally tractable approximations of the novel observer and agent value of belief expressions proposed in \eqref{eqn:vob_o_def} and \eqref{eqn:vob_a_def}, we next turn to leveraging them to perform DPP.

\subsection{VoI-based deceptive path planning} \label{subsec:formulation:voi_dpp}

% \todo{provide high-level VoI-based DPP problem formulation, then explicit LP formulation -- justify linearity (convexity?) of VoI-based objective; explain each step of how the various quantities in the LP objective are computed}

% We now move on to the problem of modeling how the observer determines which node to remove. Addressing this problem necessitates formulation of a suitable objective, which in turn requires determining observer beliefs regarding the agent's true goal, computing costs for candidate goals and candidate removal nodes, and\ldots The model that we propose builds on the observer model of Section \ref{subsec:background:observer_model} and computational machinery of Section \ref{subsec:formulation:adversarial_setting

% \todo{I would suggest, from Eqs~(\ref{eqn:occ_meas_approx}, also substituting the approximation $\text{VoB}(s)$ from (\ref{eqn:vob_approx_1}) into the RHS. As it is right now, we're condition $P(G|\zeta_{0:T})$ for the max portion.}

% \todo{propagate VoB definition changes from Section \ref{subsec:formulation:voi_defs} through the rest of the present section}

% We now propose a VoI-based approach for DPP, where the objective is based on the fundamental VoI-related definitions proposed in the previous section.
%
In the framework proposed in this section, the objective of the agent is to find a policy that generates paths minimizing either the expected observer value of belief defined in \eqref{eqn:vob_o_def} or the expected agent value of belief of \eqref{eqn:vob_a_def} while ensuring that the true goal $G^* \in \mc{G}$ is eventually reached. As we later discuss, minimizing \eqref{eqn:vob_o_def} and \eqref{eqn:vob_a_def} intuitively corresponds to encouraging the ambiguity and exaggeration notions of deception from previous works described in Section \ref{sec:background}.
%
% Despite these similarities, the experimental results presented in Section \ref{sec:experiments} demonstrate that our VoI-based formulations of deception enjoy significant advantages over existing notions in the adversarial setting considered in this paper.
%
Despite these similarities, the results presented in Section \ref{sec:experiments} indicate that our VoI formulations enjoy significant advantages in the adversarial setting, yielding shorter path lengths and improved deceptiveness over existing methods.
%
% In the remainder of the section, we derive tractable approximations of the expected value of belief objectives to be minimized, then describe tractable LP-based formulations for obtaining the desired DPP policies.

\subsubsection{Observer value of belief objective} \label{subsubsec:formulation:vob_o}
In the problem of minimizing the observer value of belief \eqref{eqn:vob_o_def}, for a fixed $T$ the agent's objective is to choose a policy $\pi$ that minimizes the expected value of \eqref{eqn:vob_approx_0} under $\pi$:
\begin{align}
    \mathbb{E}_{\zeta_{0:T} \sim \pi} & \left[ \text{VoB}^o(\zeta_{0:T}) \right] = \int_{\mc{T}_{0:T}} \text{VoB}^o(\zeta_{0:T}) \ P(\zeta_{0:T} | \pi) d \zeta_{0:T} \nonumber \\
    &= \int_{\mc{T}_{0:T}} \max_{i \in \mc{I}} \sum_{G \in \mc{G}} P(G | \zeta_{0:T}) J(\bar{\pi}^i_G) \ P(\zeta_{0:T} | \pi) d \zeta_{0:T}, \label{eqn:expected_vob_0}
\end{align}
where $\mc{T}_{0:T} = \{ \zeta_{0:T} \ | \ \zeta_{0:T} \sim \pi \}$ is the set of all possible trajectories generated by $\pi$ in MDP $\mc{M}$ up to time $T$, and $P(\zeta_{0:T} | \pi) = \Pi_{t=0}^{T-1} p(s_{t+1} | s_t, a_t) \pi(a_t | s_t)$ is the probability of $\zeta_{0:T}$ under $\pi$ in $\mc{M}$.
Recalling the approximation \eqref{eqn:belief_model}, we have that equation \eqref{eqn:expected_vob_0} is
\begin{align} \label{eqn:expected_vob_0a}
    &\approx \int_{\mc{T}_{0:T}} \max_{i \in \mc{I}} \sum_{G \in \mc{G}} \widehat{P}(G | s_T) J(\bar{\pi}^i_G) \ P(\zeta_{0:T} | \pi) d \zeta_{0:T} \\
    &= \sum_{s \in \mc{S}} \int_{ \substack{\mc{T}_{0:T}, \\ s_T = s} } \max_{i \in \mc{I}} \sum_{G \in \mc{G}} \widehat{P}(G | s) J(\bar{\pi}^i_G) P(\zeta_{0:T} | \pi) d \zeta_{0:T} \\
    &= \sum_{s \in \mc{S}} \max_{i \in \mc{I}} \sum_{G \in \mc{G}} \widehat{P}(G | s) J(\bar{\pi}^i_G) \int_{ \substack{\mc{T}_{0:T}, \\ s_T = s} }P(\zeta_{0:T} | \pi) d \zeta_{0:T} \\
    &\approx \sum_{s \in \mc{S}} \max_{i \in \mc{I}} \sum_{G \in \mc{G}} \widehat{P}(G | s) J(\bar{\pi}^i_G) \ d^{\pi}(s),
\end{align}
where the final approximation holds since, for large $T$,
\begin{equation}
    \int_{ \substack{\mc{T}_{0:T}, \\ s_T = s} } \ P(\zeta_{0:T} | \pi) \ d \zeta_{0:T} \approx d^{\pi}( s ) := \lim_{T \rightarrow \infty} P(s_T = s \ | \ \pi), \label{eqn:occ_meas_approx}
\end{equation}
where $d^{\pi} \in \Delta(\mc{S})$ is the occupancy measure induced by $\pi$ over $\mc{M}$, which provides the steady-state probability of being in a given state $s$ under $\pi$.
%
% \todo{add a brief discussion and references supporting the statement that \eqref{eqn:occ_meas_approx} provides a good approximation for large $T$}
%
% In addition, since $s_0$ is fixed, the observer model approximation \eqref{eqn:observer_model} can be used to replace $P(G | \zeta_{0:T})$ with a tractable approximation depending only on the final element $s_T$ of the trajectory $\zeta_{0:T}$.
%
Taken together, these approximations suggest the following approximation of objective \eqref{eqn:expected_vob_0}:
\begin{align}
    \mathbb{E}_{s \sim d^{\pi}} &\left[ \text{VoB}^o(s) \right] = \sum_{s \in \mc{S}} \max_{i \in \mc{I}} \sum_{G \in \mc{G}} P(G | \zeta_{0:T}) J(\bar{\pi}^i_G) \ d^{\pi}(s) \label{eqn:expected_vob_1} \\
    &= \sum_{s \in \mc{S}, a \in \mc{A}} \max_{i \in \mc{I}} \sum_{G \in \mc{G}} P(G | \zeta_{0:T}) J(\bar{\pi}^i_G) \ \lambda^{\pi}(s, a) \label{eqn:expected_vob_2} \\
    &\approx \sum_{s \in \mc{S}, a \in \mc{A}} \max_{i \in \mc{I}} \sum_{G \in \mc{G}} \widehat{P}(G | s) J(\bar{\pi}^i_G) \ \lambda^{\pi}(s, a), \label{eqn:expected_vob_3}
\end{align}
where we define $\text{VoB}^o(s) = \max_{i \in \mc{I}} \sum_{G \in \mc{G}} \widehat{P}(G | s) J(\bar{\pi}^i_G)$, and where we use the fact that the state-action occupancy measure $\lambda^{\pi}$ induced by $\pi$ over $\mc{M}$ satisfies $\lambda^{\pi}(s, a) = d^{\pi}(s) \pi(a | s)$ to rewrite \eqref{eqn:expected_vob_1} as \eqref{eqn:expected_vob_2}.
%
% \todo{add explanation of why this intuitively corresponds to ambiguity}

To see why this objective corresponds to ambiguity in the adversarial setting, notice that a policy that minimizes \eqref{eqn:expected_vob_0} (or its approximation \eqref{eqn:expected_vob_3}) shifts the probability mass of the observer's belief distribution away from candidate goals corresponding to high-cost interventions and onto candidate goals corresponding to low-cost interventions. As a result, the expected maximum cost the observer is able to inflict over all possible interventions is minimized, intuitively decreasing the observer's understanding of the true effect of the possible interventions, thereby increasing the ambiguity regarding which intervention yields highest cost.

\subsubsection{Agent value of belief objective} \label{subsubsec:formulation:vob_a}
Similar to the expected observer value of belief objective described in the previous section, for the expected agent value of belief \eqref{eqn:vob_a_def} and fixed $T$ the objective of the agent is to choose a policy $\pi$ that minimizes the expected value of \eqref{eqn:vob_a_def} under $\pi$:
\begin{align}
    \mathbb{E}_{\zeta_{0:T} \sim \pi} & \left[ \text{VoB}^a(\zeta_{0:T}) \right] = \int_{\mc{T}_{0:T}} \text{VoB}^a(\zeta_{0:T}) \ P(\zeta_{0:T} | \pi) d \zeta_{0:T} \nonumber \\
    &= \int_{\mc{T}_{0:T}} J(\bar{\pi}^{i^*}_{G^*}) \ P(\zeta_{0:T} | \pi) d \zeta_{0:T}, \label{eqn:expected_vob_a_1}
\end{align}
where $J(\bar{\pi}^{i^*}_{G^*})$ is as in \eqref{eqn:vob_a_def}. Leveraging the approximation \eqref{eqn:occ_meas_approx} from Section \ref{subsubsec:formulation:vob_o} and the approximations \eqref{eqn:hat_s_def} and \eqref{eqn:vob_a_approx_0}, we have the following natural approximation of \eqref{eqn:expected_vob_a_1}:
\begin{align}
    &\mathbb{E}_{s \sim d^{\pi}} \left[ \text{VoB}^a(s) \right] = \sum_{s \in \mc{S}} J(\bar{\pi}^{i^*}_{G^*}) d^{\pi}(s) \\
    &= \hspace{-2mm} \sum_{s \in \mc{S}, a \in \mc{A}} \hspace{-2mm} J(\bar{\pi}^{i^*}_{G^*}) \lambda^{\pi}(s, a) \approx \hspace{-2mm} \sum_{s \in \mc{S}, a \in \mc{A}} \hspace{-2mm} J(\bar{\pi}^{\widehat{i^*}}_{G^*}) \lambda^{\pi}(s, a), \label{eqn:expected_vob_a_3}
\end{align}
where we set $\text{VoB}^a(s) = J(\bar{\pi}^{\widehat{i^*}}_{G^*})$ and use \eqref{eqn:vob_a_approx_0} to obtain \eqref{eqn:expected_vob_a_3}.

Notice that a policy minimizing \eqref{eqn:expected_vob_a_1} induces a belief distribution causing the observer to select an intervention with minimal cost to the agent's reaching its true goal. Such a policy generates trajectories that exaggerate the observer's belief in certain false goals, with the objective of inducing the observer to select the intervention that is most advantageous to the path planning agent.

\subsubsection{Linear programming-based solution} \label{subsubsec:formulation:lp_solution}
Taking \eqref{opt:dpp_0} as inspiration, the problem that we wish to solve is
\begin{mini}
  {\pi}{ \mathbb{E}_{\pi} \left[ \sum_{t=0}^\infty g(s_t, a_t) \right] }{}{}
  \addConstraint{ P_{\pi}( \text{Reach}(G^*) ) }{= R_{max}(G^*), }{} \label{opt:voi_dpp_0}
\end{mini}
where the reward $g$ is designed so that the objective of \eqref{opt:voi_dpp_0} corresponds to either \eqref{eqn:expected_vob_3} or \eqref{eqn:expected_vob_a_3}. To achieve this, we simply let $g(s, a) = \gamma_a^{T_{min}(s)} \text{VoB}^o(s)$ or $g(s, a) = \gamma_a^{T_{min}(s)} \text{VoB}^a(s)$, where, as in Section \ref{subsec:background:lp_solution}, $T_{min}(s) := d_{\mf{G}}(s_0, s)$ and $d_{\mf{G}}(s', s'')$ denotes the shortest path from $s'$ to $s''$ in the graph $\mf{G}$ underlying $\mc{M}$. With this $g$ in hand, we can solve the corresponding versions of \eqref{opt:dpp_1} and \eqref{opt:dpp_2} to recover an optimal policy for performing VoI-based DPP.
%
%
%
% With this definition of $g$ in hand, we can first solve the corresponding version of \eqref{opt:dpp_1}, reproduced below for convenience,
% %
% \begin{mini}
%   {\lambda \geq 0}{ \sum_{s \in \mc{S}, a \in \mc{A}} g(s, a) \lambda(s, a) }{}{}
%   %
%   \addConstraint{ \sum_{a \in \mc{A}} \lambda(s, a) - \hspace{-5mm} \sum_{s' \in \mc{S}, a \in \mc{A}} p(s' | s, a) \lambda(s', a) }{= \mathds{1}_{ \{s_0\} }(s),}{\ \forall s }{} \label{opt:voi_dpp_1}
%   %
%   \addConstraint{ \sum_{s \in \mc{S}, a \in \mc{A}} \lambda(s, a) r(s, a) }{= R_{max}(G^*)}{}{}
%   %
%   % \addConstraint{ \sum_{s \in \mc{S}, a \in \mc{A}} \lambda(s, a)}{= 1,}{}{}
% \end{mini}
% %
% followed by \eqref{opt:dpp_2} to recover an optimal policy for performing VoI-based DPP.
\begin{figure*}[htp]
    \centering
    \includegraphics[width=\textwidth]{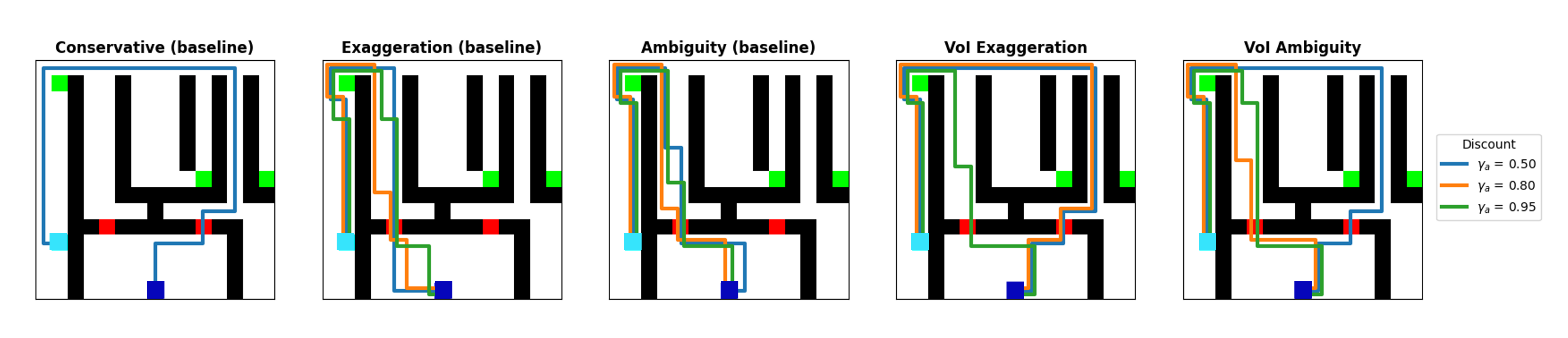}
    \caption{\small \textbf{VoI DPP enjoys flexible deceptiveness in adversarial setting.} VoI deception generates a variety of deceptive paths as $\gamma_a$ varies, while baseline methods fail by providing either shortest (Exaggeration, Ambiguity for $\gamma_a > 0.5$) or overly conservative CPP (Conservative) paths.
    %
    % For larger values of $\gamma_a$, VoI DPP generates conservative paths, while for smaller values of $\gamma_a$ the early trajectories feint towards the right-hand candidate intervention before correcting to follow the shortest path.
    %
    For small values of $\gamma_a$, VoI Exaggeration and VoI Ambiguity approximately recover the Conservative baseline.}
    \label{fig:rooms10}
\end{figure*}
\begin{figure*}[htp]
    \centering
    \includegraphics[width=\textwidth]{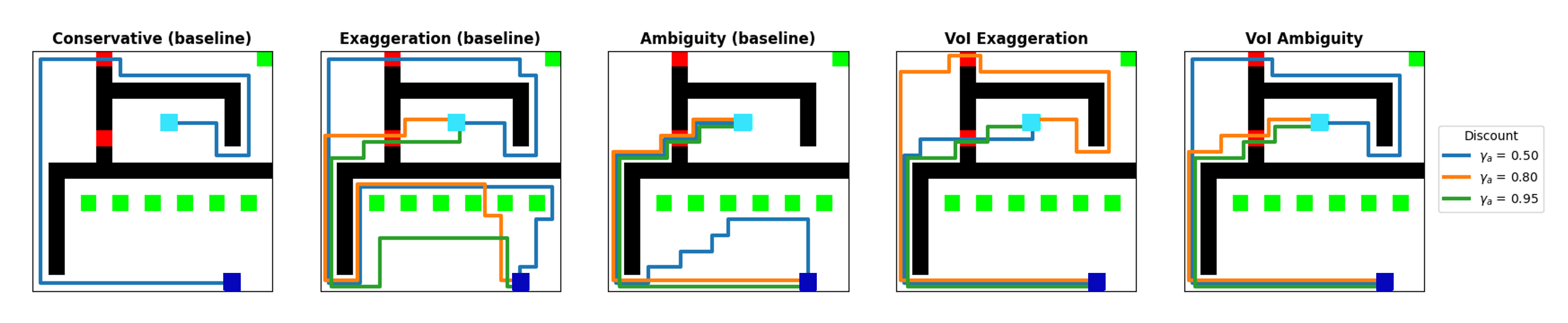}
    \caption{\small \textbf{Existing methods generate inappropriate trajectories in adversarial setting.} VoI DPP methods recognize only the top two goals are relevant, since the observer cannot affect cost of reaching remaining goals. Exaggeration (all $\gamma_a$ values) and Ambiguity ($\gamma_a = 0.5$) are distracted by irrelevant candidate goals in the lower half and generate wasteful, purposeless paths. For smaller $\gamma_a$, VoI methods approximately recover CPP paths.}
    \label{fig:rooms19}
\end{figure*}

\section{Experiments} \label{sec:experiments}

We next present experimental results comparing the methods developed in Section \ref{sec:proposed_approach} with the approaches proposed in \cite{savas2022deceptive} (see Section \ref{sec:background}) and a baseline conservative path planning (CPP) approach guaranteed to reach the true goal under worst-case observer interventions on two illustrative adversarial gridworld problems.\footnote{We derive CPP by first formulating the model proposed in Section \ref{sec:formulation} as an extensive form game where the goal of the agent is to minimize its path costs under worst-case adversarial interventions, then applying minimax value iteration to obtain a corresponding policy. See Appendix \ref{app:cpp} for details.} These experiments illustrate two main points: VoI DPP provides flexible deception in the adversarial problem setting we consider, while the passive-observer methods of \cite{savas2022deceptive} are ill-suited to adversarial problems and the CPP method lacks flexibility; under actual observer interventions, VoI DPP outperforms passive-observer and CPP methods when interventions are made during the \textit{critical deception window} (CDW) defined below.
%
% \textcolor{red}{under actual observer interventions, VoI-based deception empirically leads to lower-cost paths compared with the other methods across a variety of problems.}

% \subsection{Experiment Setup} \label{subsec:experiments:setup}
\textbf{Experiment setup.} We considered the problems depicted in Figures \ref{fig:rooms10} and \ref{fig:rooms19}. Agent start state is depicted in blue, decoy goals are green, dark grey squares represented obstacles, and the agent's true goal is light blue. The observer threatens to intervene by placing an obstacle at one of the candidate intervention points depicted in red. The objective of the observer is to select the candidate intervention that causes the greatest increase in the agent's navigation costs. This setting is naturally captured by the MDP-based formulation of Section \ref{sec:formulation}: the underlying graph consists of nodes for each of the accessible squares and edges are determined by adjacency between the corresponding squares in the gridworld. For interpretability, in the problems we consider we stipulate that every edge has the same fixed cost. % As described in Section \ref{subsec:formulation:adversarial_setting}, the observer assumes the agent is trying to minimize its costs plus an entropy regularization term, so this reduces to minimizing path length in the entropy-regularized path planning problem \eqref{opt:obs_model}.

\subsection{Qualitative path comparisons} \label{subsec:experiments:qualitative}

\begin{figure*}[htp]
    \centering
    \includegraphics[width=\textwidth]{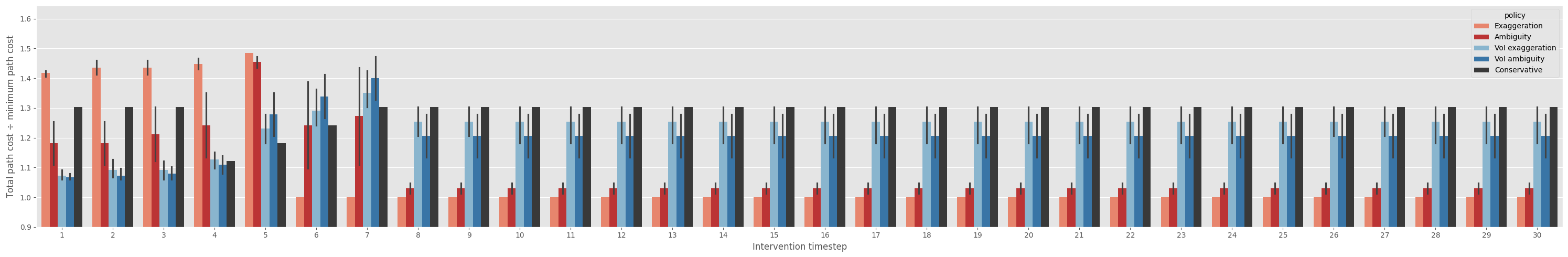}
    \caption{ \small
    \textbf{Performance comparison under adversarial interventions on Fig. \ref{fig:rooms10} gridworld.}
    %
    % Plot shows ratio of total path length to shortest path length.
    %
    Agents deceive up until intervention time, then follow shortest paths afterwards. Interventions are selected according to observer belief at time of intervention.
    Plots show ratio of total path length to the shortest path from the start state to the true goal: a value of $1.0$ corresponds to a shortest path and minimum possible value, while larger values capture the excess cost being incurred for behaving deceptively (for DPP) or conservatively (for CPP).
    Total costs are aggregated over 10 values of $\gamma_a$, box and whiskers show mean and standard deviation.
    The \textit{critical deception window} is from timesteps 1 to 5, during which VoI-based deception is particularly effective, outperforming pass-observer methods and outperforming or remaining competitive with CPP.
    }
    \label{fig:rooms10_critical_window}
\end{figure*}
%
% \vspace{-2cm}
%
\begin{figure*}[htp]
    \centering
    \includegraphics[width=\textwidth]{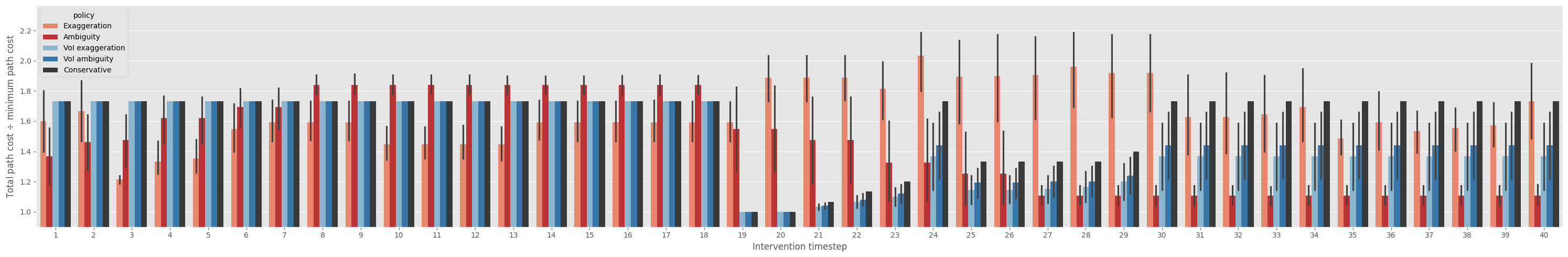}
    \caption{\small \textbf{Performance comparison under adversarial interventions on Fig. \ref{fig:rooms19} gridworld.}
    Agents deceive up until intervention time, then follow shortest paths afterwards.
    Interventions are selected according to observer belief at time of intervention.
    Plots show ratio of total path length to the shortest path from the start state to the true goal.
    %
    % a value of $1.0$ corresponds to a shortest path and minimum possible value, while larger values capture the excess cost being incurred for behaving deceptively (for DPP) or conservatively (for CPP).
    %
    Total costs are aggregated over 10 values of $\gamma_a$, box and whiskers show mean and standard deviation.
    The \textit{critical deception window} in this problem is from timesteps 19 to 30.
    %
    % due to the increased distance of the intervention points from the start state in Fig. \ref{fig:rooms19}.
    %
    Throughout this time VoI deception significantly outperforms classical exaggeration methods and either outperforms or remains competitive with CPP. VoI approaches tend to outperform classical ambiguity through timestep 26, but are subsequently outperformed by it.}
    \label{fig:rooms19_critical_window}
\end{figure*}

Figures \ref{fig:rooms10} and \ref{fig:rooms19} illustrate three main points: VoI DPP methods generate a variety of deceptive behaviors in adversarial settings, existing passive-observer methods are ill-suited to adversarial settings, and the CPP method lacks flexibility and is recovered by VoI DPP methods as a special case.

\textbf{VoI-based DPP flexibility.} Figures \ref{fig:rooms10} and \ref{fig:rooms19} indicate that VoI-based deception leads to a range of deceptive path types as $\gamma_a$ varies: the longest, most deceptive paths occur for smaller values of the discount $\gamma_a > 0$, while these approach shortest paths as $\gamma_a \rightarrow 1$. Small values of $\gamma_a$ lead to long, highly deceptive paths in all cases. As Figure \ref{fig:rooms10} illustrates, however, even for larger $\gamma_a$ values ($\gamma_a = 0.95$ for VoI Exaggeration and $\gamma_a = 0.8, 0.95$ for VoI Ambiguity) both VoI metrics remain deceptive by feinting toward the right-hand intervention early on. Notice that the right-hand intervention is suboptimal for the observer in Figure \ref{fig:rooms10}, since it leaves the shortest path to the true goal open to the agent.

\textbf{Drawbacks of passive-observer methods.} Figures \ref{fig:rooms10} and \ref{fig:rooms19} illustrate the unsuitability of the passive-observer methods proposed in \cite{savas2022deceptive} for the adversarial setting. Figure \ref{fig:rooms10} illustrates that these methods can fail to generate deceptive trajectories, instead returning shortest paths even as $\gamma_a$ varies (with the exception of Ambiguity with $\gamma_a = 0.5$). Figure \ref{fig:rooms19} demonstrates that passive-observer methods also become ``distracted'' in the adversarial setting by candidate goals that are irrelevant in the sense that the observer is unable to intervene to affect the cost to the agent of reaching them. The result is that the rewards \eqref{eqn:classical_exaggeration} and \eqref{eqn:classical_ambiguity} often generate detours that accrue additional cost without impacting observer interventions. While this is expected, it nonetheless highlights their inappropriateness in the adversarial setting.

\textbf{Conservative method as VoI DPP special case.} In Figures \ref{fig:rooms10} and \ref{fig:rooms19}, the paths generated by CPP policies are clearly deceptive, since their path costs are highly suboptimal for reaching the true goal in the absence of interventions. This approach is effective when deception is the first priority and path cost is secondary, but inflexible when varying trade-offs between deceptiveness and path cost are required. As shown in Section \ref{subsec:experiments:quantitative}, addressing this trade-off enables VoI DPP policies to leverage deception to achieve lower total path costs under interventions. Finally, it is important to note that, for low values of $\gamma_a$, both VoI metrics approximately recover the conservative paths as a special case. In addition to supporting the generality of our methods, this suggests a connection between minimax value iteration and VoI-based deception that merits further exploration.

\subsection{Comparison under interventions} \label{subsec:experiments:quantitative}

For Figures \ref{fig:rooms10_critical_window} and \ref{fig:rooms19_critical_window}, we compared the performance of passive-observer, CPP, and VoI policies under actual interventions on the problems pictured in Figures \ref{fig:rooms10} and \ref{fig:rooms19}, respectively. In the setting we considered, a given path planning agent follows its policy up until intervention time, the observer selects the optimal intervention corresponding to its belief distribution based on the agent trajectory up to that time (see \eqref{eqn:expected_vob_0}), then the agent proceeds to its true goal via the minimum-cost path in the post-intervention problem.
%
% We considered intervention times between $t = 1$ and $t = 50$ for all policies, and for each of the DPP methods we evaluated 10 policies, one for each $\gamma_a \in [0.5, 0.55, \ldots, 0.9, 0.95]$. The figures report the ratio of the total path cost accumulated to the minimum cost (i.e., shortest) path from the start state to the true goal: a value of $1.0$ thus corresponds to a shortest path and minimum possible value, while larger values capture the excess cost being incurred for behaving deceptively (for DPP) or conservatively (for CPP). For the DPP methods, the values reported are averaged over the performances achieved over the 10 policies evaluated.

Due to the locations of the start state, interventions, and goals within the structure of the underlying problem, there is a window of time where both intervention and deception have a particularly strong affect on the total path cost achieved by the agent. We refer to this window as the \textit{critical deception window} (CDW) (see Figures \ref{fig:rooms10_critical_window} and \ref{fig:rooms19_critical_window}).
Within the CDW the observer's belief distribution is highly sensitive to changes in the agent's location, while the agent's total path cost is highly sensitive to interventions. Outside this window both the observer's belief distribution and the agent's total path cost remain relatively static. When interventions occur during the CDW, our VoI policies tend to outperform the other methods, highlighting that the VoI deception metrics \eqref{eqn:vob_o_def} and \eqref{eqn:vob_a_def} capture the impact of beliefs on intervention selection and VoI DPP therefore leverages deception to achieve lower-cost paths when interventions occur during the CDW.
%
% our VoI-based policies tend to significantly outperform the passive-observer policies and either outperform or remain competitive with CPP policies. This highlights the fact that our VoI deception metrics \eqref{eqn:vob_o_def}, \eqref{eqn:vob_a_def} are explicitly designed to evaluate the impact of beliefs on intervention selection, and that VoI DPP therefore leads to policies that leverage deception to achieve lower-cost paths when interventions occur during the CDW.
%
Outside this window, on the other hand, performance is highly problem-dependent: VoI methods outperform or remain competitive with CPP on both problems, yet passive-observer methods outperform in Figure \ref{fig:rooms10_critical_window} and performance outside the CDW is mixed on Figure \ref{fig:rooms19_critical_window}. Overall, the superior performance of VoI DPP within the CDW underscores the importance of our proposed methods, but the significance of the CDW within the context of DPP merits additional exploration.

\section{Conclusion}

In this work we developed value of information-based approaches to the problem of deceptive path planning under adversarial interventions. To achieve this, we proposed a novel model for DPP in the adversarial settings, proposed VoI-based deception metrics that quantify the impact of deceptive behaviors via the informational value they furnish to the observer, and derived efficient, LP-based solution procedures to obtain policies for performing VoI DPP, and provided experimental results validating their effectiveness on illustrative gridworld problems. While this work contributes a solid foundation for DPP in adversarial settings and promising empirical results, more in-depth experimental validation and exploration of the significance of critical deception windows are needed, providing important directions for future work.

%%%%%%%%%%%%%%%%%%%%%%%%%%%%%%%%%%%%%%%%%%%%%%%%%%%%%%%%%%%%%%%%%%%%%%%%%%%%%%%%
\section*{ACKNOWLEDGMENTS}

This work was supported in part by the U.S. Army Research Office grant W911NF-23-1-0317, Army Research Laboratory grant W911NF-25-2-0021, and Office of Naval Research grant N00014-24-1-2432.

%%%%%%%%%%%%%%%%%%%%%%%%%%%%%%%%%%%%%%%%%%%%%%%%%%%%%%%%%%%%%%%%%%%%%%%%%%%%%%%%

\bibliographystyle{IEEEtran}
\bibliography{refs}

\appendix
\section{Appendix}

\subsection{Extended Related Works} \label{app:related_works}

% \textbf{Goal recognition and obfuscation.} The literature on goal recognition and goal obfuscation is closely related to the deceptive planning setting considered in our work. In the goal obfuscation literature, \cite{kulkarni2018resource, bernardini2020optimization} considered the problem of generating paths over a graph that conceal the agent's true objective from a passive, external observer. The subsequent paper \cite{takahashi2024transit} addressed the transit obfuscation problem, an extension of goal obfuscation to the problem of concealing an intermediate ``transit'' point that must be visited en route between the start and goal states from a passive observer. Adjacent to this, goal recognition works including \cite{masters2021extended, price2023domain} address the problem of modeling observer inference of the agent's true goal. While \cite{masters2021extended} focuses primarily on observer modeling, \cite{price2023domain} proposes corresponding deception metrics and formulates deceptive planning approaches. We emphasize that the goal obfuscation and recognition works discussed above focus on the passive-observer setting.

\textbf{Active-observer deception.} While previous methods for DPP are restricted to the passive-observer setting, there exists previous work on deceptive planning under adversarial interventions and deception in games. The paper \cite{cates2023planning} considered the problem of deceptive planning for a defender attempting to guide an attacker into a set of trap states while concealing its intentions from the attacker. There are two key differences between the setting of \cite{cates2023planning} and that considered in our paper: (i) our approach addresses the attacker-/agent-perspective problem of previous works \cite{ornik2018deception, savas2022deceptive, chen2024deceptive, fatemi2024deceptive}, while \cite{cates2023planning} considers the distinct defender-/observer-perspective deceptive planning problem; (ii) the notion of deception considered in \cite{cates2023planning} is based on a binary hypothesis test of whether the attacker realizes it is being deceived, while in our work we develop general VoI deception metrics capturing the value of arbitrary observer belief distributions.

There have been several works on deception in games that incorporate active observers, including \cite{ma2023optimal, shi2023quantitative, kulkarni2024integrated, rostobaya2023deception, karabag2024identity}. In both \cite{ma2023optimal} and \cite{kulkarni2024integrated}, the problem of allocating defensive resources against an attacker operating over a graph is considered, where the adversary's best response is explicitly considered. The work \cite{shi2023quantitative} addressed a competitive goal-reaching game, where one agent exploits partial observability of its actions to improve its payoff. In \cite{karabag2024identity}, identity concealment games are proposed and studied, and an offline learning approach to obtaining an approximately optimal policy against a suboptimal adversary is provided. Finally, the work \cite{rostobaya2023deception} considered a particular formulation of the DPP problem where the observer can modify the final payoff of the agent depending on its agent trajectory observations. It is important to note that, though equilibria are successfully characterized in the foregoing works, computation of equilibrium policies remains a challenge. This contrasts with the adversarial setting considered in our work, which permits computationally tractable, linear programming-based solution procedures at the expense of flexibility in our modeling of the observer.

\textbf{Value of information.} The notion of the value of information (VoI) originates from early work in statistical decision theory \cite{howard1966information,raiffa2000applied} and comparison of experiments \cite{blackwell1953equivalent,cremer1982simple}.
%
% In this classical formulation, information is understood as an experiment performed by a decision-maker, in order to reduce uncertainty for a stochastic optimization problem depending on an unknown world state.
%
% Both the single- and multi-agent concept of VoI can be subsumed under the value of `information structures'  for decision-makers, see e.g. \cite{pkeski2008comparison}.
%
While we are not aware of any works explicitly using VoI for deception, \cite{greenberg1982role} noted that misleading manipulation of an opponent's information to one's advantage is considered deceptive -- for zero-sum games, this is equivalent minimizing the opponent's VoI. It has been shown in both static \cite{hespanha2000deception} and dynamic \cite{rostobaya2023deception} games that deceptive signaling can be Nash optimal, without assuming a na\"{i}ve observer or an exogenous reward for deception as in previous DPP works \cite{ornik2018deception,savas2022deceptive,chen2024deceptive,karabag2021deception}. While our approach does assume the na\"{i}ve observer of these works, we dispense with an exogenous deception reward, and instead take a step toward the game-theoretic perspective by allowing the observer to act with respect to their subjective utility. This work offers a middle ground where meaningful deception on the part of the agent occurs within an adversarial scenario, while still offering a general, computationally tractable control-based approach to DPP.

\textbf{Conservative path planning.} Path planning under interventions can be performed via conservative approaches that minimize the worst-case value, i.e., total discounted cost. These approaches include using a robust MDP~\cite{NE:05} where the transition probability intervals capture all feasible MDPs under interventions. However, this approach may result in infeasibility as it allows all interventions to take place simultaneously. Alternatively, planning under interventions can be modeled as an extensive form game~\cite{HART199219}. The game states represent the state in the MDP and the agent's information set on whether each intervention is performed. The observer aims to maximize the total discounted cost of the agent by modifying the MDP and, accordingly, the information set, and the agent aims to minimize the cost by transitioning in the MDP and, thereby, acquiring information.  While conservative path planning approaches minimize the worst-case value, they are agnostic to the observer's beliefs and may result in overly conservative plans, especially if the 
agent can influence the beliefs and interventions of the observer. 
Our approach aims to minimize the cost of the agent while taking possible interventions into account and influencing them.

\subsection{Conservative path planning approach} \label{app:cpp}

%%%%%%%%%%%%% REVISED VERSION
While our goal in this work is to develop path planning methods that leverage deception to achieve path costs that are better than worst-case, it is important to have a conservative method that is guaranteed to reach the true goal under worst-case observer interventions. In this subsection, we provide such a method by first formulating the model proposed in Section \ref{sec:formulation} as an extensive form game where the goal of the agent is to conservatively minimize its path costs under worst-case adversarial interventions, then applying minimax value iteration to obtain a corresponding policy.
%
% An extensive form game formulation between the agent and the observer can conservatively optimize (i.e., minimize the worst-case) the agent’s objective function under an intervention.

In the game, the agent has an information set $\lbrace 0, 1, ? \rbrace^{k}$, where the $i$th element is $?$ if the agent does not know whether the $i$th intervention is performed, $0$ if the agent knows that the $i$th intervention is not performed, and $1$ if the agent knows that the $i$th intervention is performed, and where the number of $1$s is at most $1$ since at most $1$ intervention can be performed. The state $\bar{s} \in \mathcal{S} \times \lbrace 0, 1, ? \rbrace^{k}$ represents the state of the agent in the MDP and the current knowledge of the agent about the interventions. At state (s, I), the transition dynamics of the MDP $\mc{M}$ follows the induced MDP $\bar{M}_{i}$ if the $i$th element of $I$ is $1$, i.e., the $i$th intervention is performed, and follows the original dynamics of $M$ if $I \in \lbrace 0, ?\rbrace^{k}$, i.e., none of the interventions has been performed yet. 
%The function $\intervene: \mathcal{S} \to 2^{\lbrace 1, \ldots, k\rbrace}$ represents the possible interventions that can be made when the agent is at state $s$. 
If the observer decides to perform intervention $i$ at state $(s, I)$, the agent moves to state $(s, \tilde{I})$, where $\tilde{I}_{i} = 1$ and $\tilde{I}_{j} =0$ for all $j \neq i$. If the observer reveals that intervention $i$ has not yet been performed, the agent moves to $(s, \tilde{I})$, where $\tilde{I}_{i} = 0$ and $\tilde{I}_{j} =I_{j}$, for all $j \neq i$. Let $Succ(s, I)$ denote all possible information sets that can be transitioned to from $(s, I)$.

From \cite{littman1994markov}, the optimal policy of the agent depends on the MDP state and the information set, and the value function $V^{*}$ satisfies
\begin{align}
    V^*(s, I) &= \max_{\tilde{I} \in Succ(I)} V^{*}(s,\tilde{I}), \text{ for all } (s, I) \in \lbrace 0, 1, ? \rbrace^{k}, \label{eqn:pmvi_1} \\
    V^{*}(s, I) &= \min_{a \in \mathcal{A}(s)} c(s, a) + \gamma \sum_{s’ \in \mathcal{S}}p(s’|s,a) V^{*}(s',I), \label{eqn:pmvi_2}
\end{align}
%
% $$V^*(s, I) = \max_{\tilde{I} \in Succ(I)} V^{*}(s,\tilde{I}), \text{ for all } (s, I) \in \lbrace 0, 1, ? \rbrace^{k} ,$$ and
% $$V^{*}(s, I) = \min_{a \in \mathcal{A}(s)} c(s, a) + \gamma \sum_{s’ \in \mathcal{S}}p(s’|s,a) V^{*}(s',I)$$
%
for all $(s, I)$ such that $I  = \arg\max_{\tilde{I} \in Succ(I)} V^{*}(s,\tilde{I})$.
In words, equations \eqref{eqn:pmvi_1}, \eqref{eqn:pmvi_2} ensure that, if the observer does not perform an intervention or reveal information, the agent moves optimally according to the current information set. If the observer performs an intervention or reveal information at $(s,I)$, then the state $s$ acts as a terminal state for the information set $I$ with the final cost of $V^{*}(s,I) = \max_{\tilde{I} \in Succ(I)} V^{*}(s,\tilde{I})$. Assigning arbitrarily large $c(s,a)$ to $s \in \mathcal{G} \setminus \lbrace G^{*} \rbrace$ ensures that the output policy reaches $G^{*}$ with the maximum probability. 
One can solve the set of optimality equations \eqref{eqn:pmvi_1}, \eqref{eqn:pmvi_2} inductively by starting with $I \in \lbrace 0, 1 \rbrace^{k}$, i.e., there are no unknown intervention locations, via minimax value iteration~\cite{littman1994markov}. Having computed $V^{*}$ for states with $n$ unknown interventions, one can compute the values for $n+1$ interventions by considering the states with $n$ unknown interventions as terminal states with the final costs computed in the previous induction steps. This procedure has $\mathcal{O}(2^k)$ computational complexity (ignoring the factors related to the size of $\mathcal{M}$) due to the possible information set configurations.

As illustrated in Section \ref{sec:experiments}, the CPP scheme detailed above yields a conservative yet effective worst-case approach for the path planning problem. Interestingly, the VoI-based DPP methods developed in the following section approximately recover the CPP method as a special case (see Figures \ref{fig:rooms10}, \ref{fig:rooms19}), despite the fact that the computational complexity of the latter methods scales linearly in $k$.
%(again ignoring factors related to the size of $\mathcal{M}$).

\end{document}